\documentclass[letterpaper]{article} 
\usepackage[]{aaai23}  
\usepackage{times}  
\usepackage{helvet}  
\usepackage{courier}  
\usepackage[hyphens]{url}  
\usepackage{graphicx} 
\usepackage{natbib}  
\usepackage{caption} 
\usepackage{multirow}
\usepackage{booktabs}
\usepackage{amssymb}

\usepackage{algorithm}
\usepackage{algorithmic}
\usepackage{marvosym}

\usepackage{newfloat}
\usepackage{listings}
\DeclareCaptionStyle{ruled}{labelfont=normalfont,labelsep=colon,strut=off} 
\floatstyle{ruled}
\newfloat{listing}{tb}{lst}{}
\floatname{listing}{Listing}

\nocopyright

\title{STS: Surround-view Temporal Stereo for Multi-view 3D Detection}

\author{
    Zengran Wang,
    Chen Min,
    Zheng Ge\textsuperscript{\Letter},
    Yinhao Li,\\
    Zeming Li,
    Hongyu Yang,
    Di Huang
}
\affiliations{
    MEGVII Technology, Peking University, Beihang University\\
    wangzengran@megvii.com, minchen@stu.pku.edu.cn,\\
    \{gezheng, liyinhao,  lizeming\}@megvii.com, \\
    \{hongyuyang, dhuang\}@buaa.edu.cn
     
}

\usepackage{bibentry}

\begin{document}

\maketitle

\begin{abstract}

Learning accurate depth is essential to multi-view 3D object detection. Recent approaches mainly learn depth from monocular images, which confront inherent difficulties due to the ill-posed nature of monocular depth learning. Instead of using a sole monocular depth method, in this work, we propose a novel Surround-view Temporal Stereo (STS) technique that leverages the geometry correspondence between frames across time to facilitate accurate depth learning. Specifically, we regard the field of views from all cameras around the ego vehicle as a unified view, namely surround-view, and conduct temporal stereo matching on it. The resulting geometrical correspondence between different frames from STS is utilized and combined with the monocular depth to yield final depth prediction. Comprehensive experiments on nuScenes show that STS greatly boosts 3D detection ability, notably for medium and long distance objects. On BEVDepth with ResNet-50 backbone, STS improves mAP and NDS by 2.6\% and 1.4\%, respectively. Consistent improvements are observed when using a larger backbone and a larger image resolution, demonstrating its effectiveness.
\end{abstract}


\section{Introduction}
Multi-view 3D detection in Bird's-Eye-View (BEV) is going through a rapid development recently, since it is a feasible alternative to LiDAR-based solutions with much lower cost for robotics and autonomous driving systems. However, perceiving depth from RGB images has long been a challenging problem due to the lack of sufficient information. Improving depth for camera 3D detection thus raises more and more attention from the research community~\cite{bevdepth}.

\begin{figure*}[t]
\centering
\includegraphics[width=1.0\textwidth]{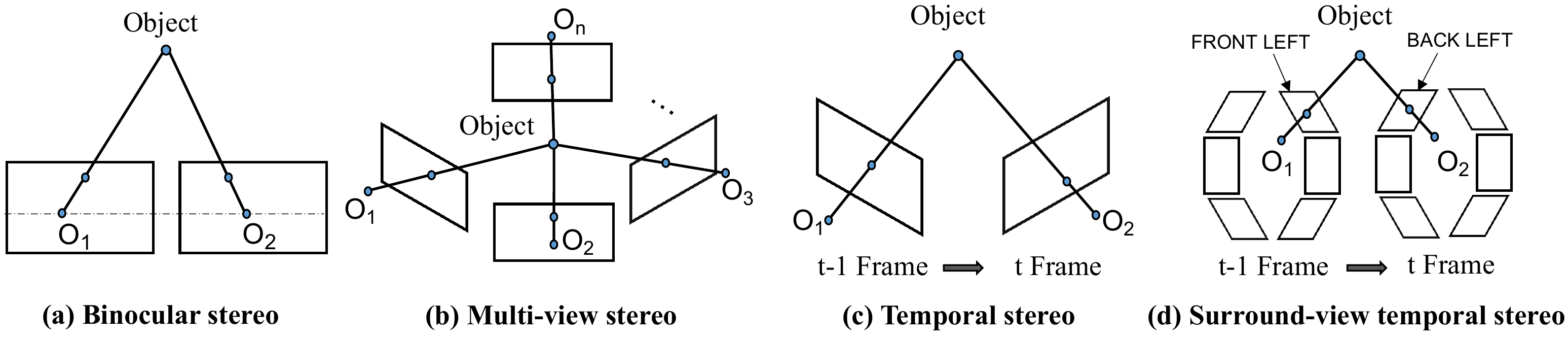}
\caption{Comparison of different settings for stereo-based depth estimation from 2D RGB images to learn the absolute scale of the world.}
\label{fig:compare}
\end{figure*}

\begin{figure}[!t]
\centering
\includegraphics[width=0.45\textwidth]{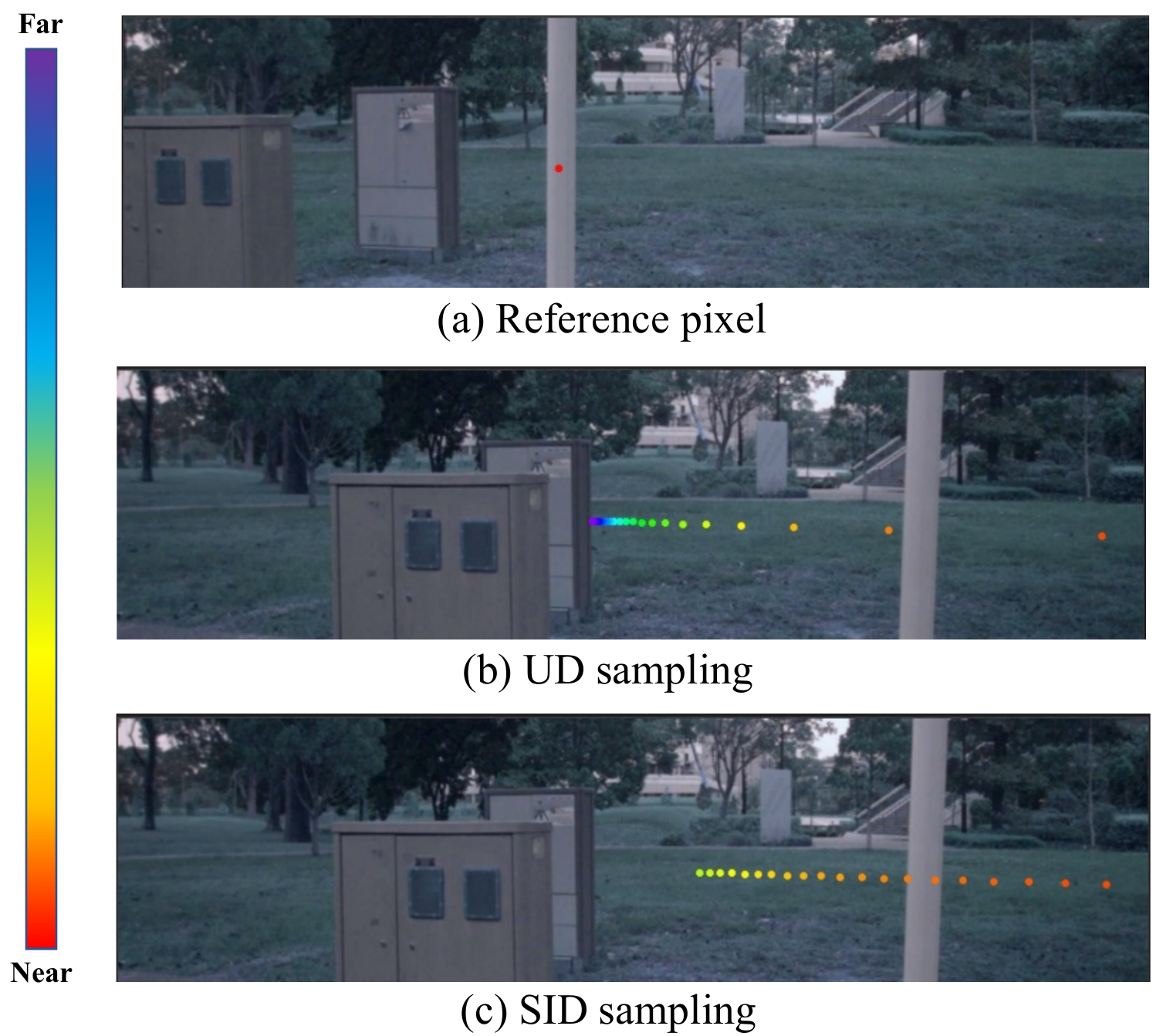} 
\caption{Visualization of projected sampling points on source images using different depth sampling strategies. UD represents traditional uniform depth sampling, SID represents Spacing-Increasing Discretization.}
\label{fig:sid_ud_com}
\end{figure}

Depth estimation, as a fundamental computer vision task, has been thoroughly explored by many means, including monocular paradigm~\cite{7298660, make3d, dorn}, binocular stereo matching~\cite{gcnet, psmnet, ganet}, multi-view stereo~\cite{mvsnet, aa-rmvsnet, magnet} and temporal stereo~\cite{dfm2} (also called depth from motion). The most common approach among them is monocular depth estimation, which directly learns depth from a raw RGB image. It estimates depth based on visual cues, such as shadows, ground plane, and the vertical position of objects~\cite{dijk2019neural}. However, learning depth from a single RGB image is ill-posed. To change its ill-posed nature, at least two images are needed. This leads us to stereo-based methods.

Stereo-based methods estimate depth based on pixel-wise correspondence between multiple views. The popular stereo depth estimation system is binocular stereo matching (Figure ~\ref{fig:compare} (a)) which needs two fronto-parallel cameras on the same plane. Another stereo-based depth estimation method is multi-view stereo (MVS, Figure~\ref{fig:compare} (b)), which uses multiple cameras (typically $\textgreater$ 2) and the relative position between cameras is given. MVS is broadly used in 3D reconstruction. The last paradigm named temporal stereo (also called depth from motion, Figure ~\ref{fig:compare} (c)) estimates depth on a sequence of images giving that the ego-motion between images is known. Despite the great development of depth estimation methods, in multi-view 3D detection, the common method is still based on monocular image and the potential of stereo-based methods is barely studied.

In multi-view 3D object detection, the field of view (FoV) of different cameras does not necessarily overlap, making it infeasible to leverage traditional stereo-based techniques. However, the overlap over time is normal. Particularly, this overlap may happen across cameras. For example, an object in ``FRONT\_LEFT'' camera at timestamp $t-1$ may appear in ``BACK\_LEFT'' camera at timestamp $t$, giving us an opportunity to take advantage of temporal stereo. Therefore, based on the specificity of our task, we explore a new paradigm of temporal stereo, namely Surround-view\footnote{An FoV of a surround-view covers FoVs from all cameras.} Temporal Stereo (STS), to facilitate accurate depth learning in multi-view 3D detection (Figure~\ref{fig:compare} (d)). Specifically, STS first generates $C_D$ depth hypotheses for each spatial location in the reference frame, and then warps their corresponding features in source frames from the history to the reference frame using the differentiable homography warping. Different from existing temporal stereo methods which only explore the correspondence within the same camera across time, STS allows the correspondence across cameras, which can make the best use of geometrical correspondence between frames. Moreover, based on the fact that a uniform hypothetical depth distribution in a reference image will sample non-uniform features in a source image (sparse in short distances but dense in long, see Figure~\ref{fig:sid_ud_com}), the features at different distances will be either over-sampled or under-sampled. We propose to leverage a non-uniform depth hypotheses, \emph{i.e.} Spacing-Increasing Discretization (SID~\cite{dorn}), to resolve this issue. This shares the same spirit with the Inverse Depth Hypotheses~\cite{xu2020learning} in MVS. Finally, because temporal stereo depth estimation struggles in texture-less regions and with moving objects, we combine the depth prediction from STS and the monocular depth module to make them compensate for each other.

We conduct extensive experiments on nuScenes~\cite{nuscenes} dataset with BEVDepth~\cite{bevdepth} as the base detector. Results show that STS greatly boosts the performance of 3D detection, notably for objects at mid-and long-range distances which can hardly be handled by monocular depth estimation. To summarize, our contributions in this work are three folds:
\begin{itemize}
    \item We bring the temporal stereo technique in multi-view 3D object detection, proposing a new stereo paradigm, namely Surround-view Temporal Stereo (STS), to facilitate accurate depth learning and 3D detection. 
    \item We conduct thorough experiments to analyze the positive effects carried by STS, as well as their different roles in the overall algorithm.
    \item Our method achieves new state-of-the-art in multi-view 3D object detection.
\end{itemize}

\section{Related Work}

\subsection{Multi-camera 3D Object Detection}

Recently, detecting objects from multi-cameras has been popular due to its low cost compared to LiDAR-based methods. DETR3D~\cite{detr3d} and PETR~\cite{petr} set object queries in 3D space and interact with
the multi-view image in the transformer decoder. BEVDet~\cite{bevdet} and BEVDet4D~\cite{bevdet4d} follow LSS~\cite{lss} which first lifts each image individually into a frustum of features for each camera and then splats all frustums into a unified bird’s-eye-view representation, and applies the detection head like CenterPoint~\cite{centerpoint} to get final 3D detection results. BEVFormer~\cite{bevformer} introduces the spatial cross-attention to
aggregate image features and the temporal self-attention to fuse the historical BEV features. In the above paradigm, depth estimation matters a lot to obtain high-quality BEV features. BEVDepth~\cite{bevdepth}  attempts to learn the trust-worthy depth and then projects image features to BEV space. However, depth estimation still brings quantitative error for the detection head. Thus we propose to improve multi-view 3D object detection with accurate depth estimation.

\subsection{Monocular Depth Estimation}

Monocular depth estimation is important for scene understanding and 3D reconstruction.
Estimating monocular dense depth is an ill-posed problem. To overcome this problem, many works try to extract monocular cues, 
such as texture variations, brightness and color, occlusion boundaries, surface, object sizes, and locations. Early works make use of Markov Random Field (MRF) and its variants with handcrafted features to estimate depth~\cite{7298660, make3d, Liu2014DiscreteContinuousDE}. 
Recently, learning-based approaches achieve state-of-art in monocular depth estimation tasks. DORN~\cite{dorn} designs an ordinary regression loss with a spacing-increasing discretization (SID) strategy to discretize dept. And they adopt a multi-scale network structure to extract multi-level features.
Further, Adabins~\cite{adabins} formulates this problem as a classification task by adaptively dividing the depth bins. What's more, outdoor monocular depth estimation can also rely on the ground to predict the depth, which benefits a lot for 3D object detection.
MonoGround~\cite{monoground} finds the importance of ground in depth estimation and 3D object detection. They introduce the ground plane as a prior to help depth estimation.

\subsection{Stereo Depth Estimation}
According to the geometrical relationship of the cameras, the stereo-based algorithms can be divided into: stereo matching, multi-view stereo, and temporal stereo methods. Stereo matching methods estimate the disparity between two cameras on the same plane. GCNet~\cite{gcnet} first introduces the 4D cost volume to stereo matching and uses the soft argmin operation to figure out the best matching results. Many variants of GCNet have been proposed, such as PSMNet~\cite{psmnet}, GwcNet~\cite{gwcnet}, GANet~\cite{ganet}, AANet~\cite{aanet} and ACVNet~\cite{acvnet}. Multi-view stereo extends the camera setting of stereo matching into arbitrary positions. The pioneering MVSNet~\cite{mvsnet} builds the cost volume with the differentiable homography. The following R-MVSNet~\cite{rmvsnet}, AA-RMVSNet~\cite{aa-rmvsnet}, CasMVSNet~\cite{casmvsnet}, PatchmatchNet~\cite{patchmatchnet}, Uni-MVSNet~\cite{uni-mvsnet} improve MVSNet with recurrent regularization or coarse-to-fine strategy. Temporal stereo methods estimate the depth of multi-frame images from a single moving camera. ManyDepth~\cite{manydepth}, MonoRec~\cite{monorec}, DepthFormer~\cite{depthformer}, and MaGNet~\cite{magnet} encode the information of multiple sequence images in a 4D cost volume similar to multi-view stereo. Inspired by the success of the above methods, we design the stereo-based depth estimation framework to learn high-quality BEV representation for multi-view 3D object detection.

\begin{figure*}[t]
\centering
\includegraphics [width=1.0\textwidth]{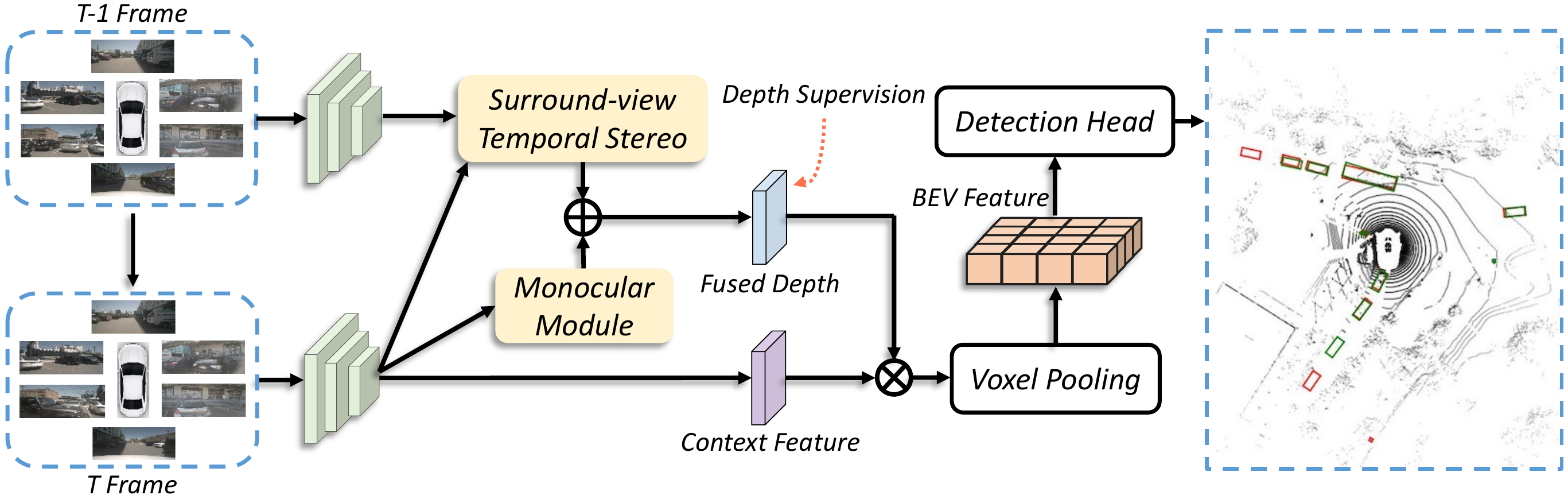}
\caption{The flowchart of our method. The image features are first lifted into a frustum of features for each camera with the depth fused from monocular depth module and STS. Then all frustums are splatted into a unified Bird’s-Eye-View representation using a pooling operation. The detection head is used to get the final detection results.}\vspace{3mm}
\label{fig:flowchart}
\end{figure*}

\begin{figure}[!t]
\centering
\includegraphics[width=0.45\textwidth]{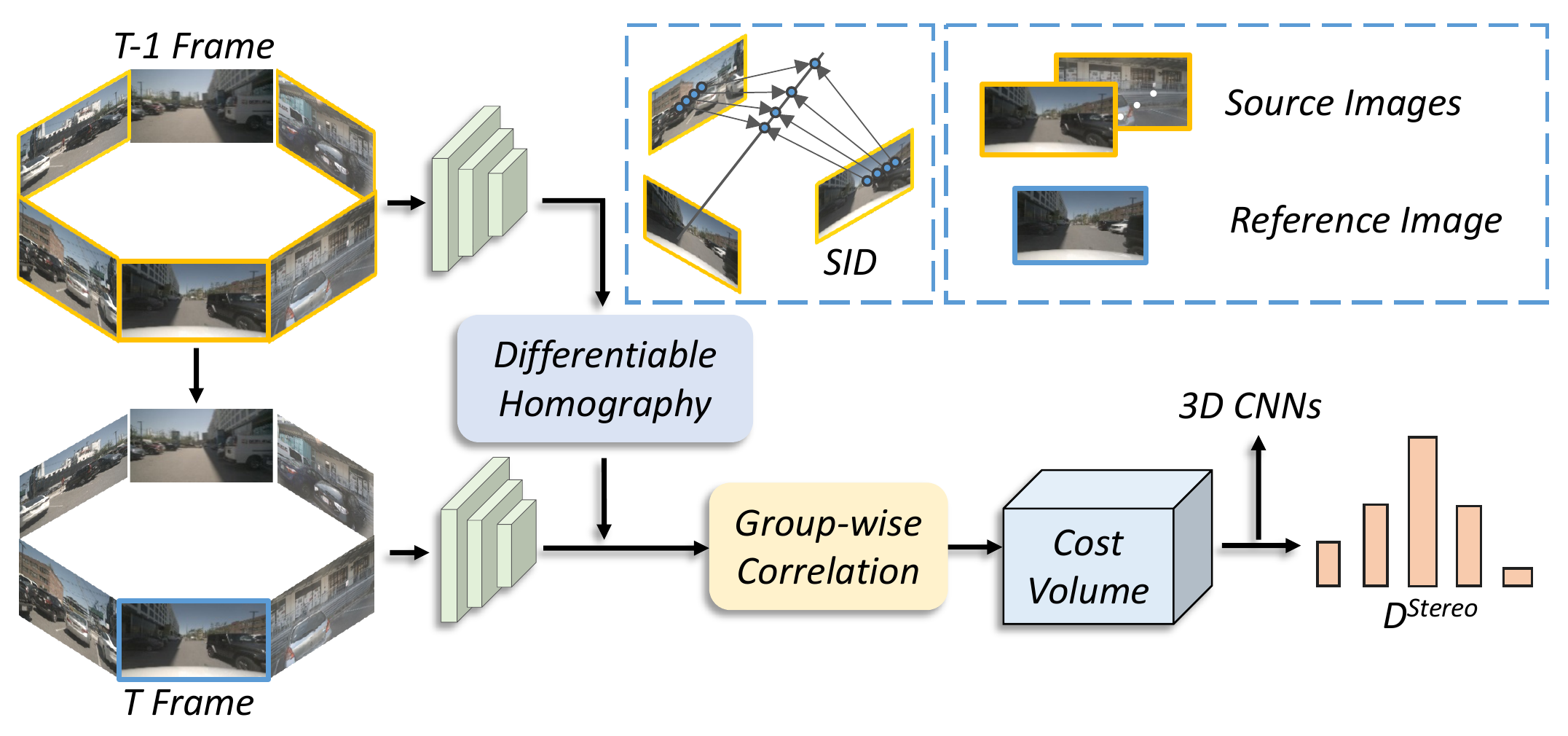} 
\caption{The detailed architecture of our proposed Surround-view Temporal Stereo (STS).}
\label{fig:sts}
\end{figure}

\section{Method} \label{method}

\subsection{Overall Architecture}

The overall architecture of our method follows the general design in multi-view 3D detection~\cite{bevdepth}. As shown in Figure~\ref{fig:flowchart}, it consists of four main stages: Image Feature Extraction, Depth Estimation, View Transformation and 3D Detection. (1) An image backbone extracts 2D features $F=\{F_{i} \in \mathbb{R}^{C_F \times \frac{H}{n} \times \frac{W}{n}}, i=1,2,...,N \}$ from $N$ views of images $I=\{I_i \in \mathbb{R}^{3 \times H\times W}, i=1,2,...,N\}$, where $H$ and $W$ denote the image's height and width. $C_F$ is the feature channel number, and $n$ stands for the down-sampling factor. (2) The Surround-view Temporal Stereo (STS) and Monocular Depth Module are leveraged to generate a fused depth $D^{pred}=\{D_i \in  \mathbb{R}^{C_D \times \frac{H}{m} \times \frac{W}{m}}, i = 1, 2, ..., N \}$, where $C_D$ is the number of sampled discrete depths, and $m$ is the down-sampling factor for depth prediction. (3) The View Transformation stage transforms 2D features $F$ onto BEV space by firstly lifting image features into a frustum of features using $D^{pred}$ and then splatting all frustums into BEV grids with a pooling operation. (4) A Detection Head predicts the class, 3D bound box location, offsets, and other attributes based on BEV features. Since this work aims at improving the quality of $D^{pred}$, in the following part, we mainly focus on the Depth Estimation stage.

\subsection{Surround-view Temporal Stereo}

\paragraph{Cost Volume Building}
The key part of temporal stereo-based depth estimation is to warp the features of the previous frame to the current frame to explore the geometry constraint across time~\cite{dfm}. However, traditional temporal methods can only explore the temporal consistency within the same camera. Our surround-view temporal stereo, on the other side, allows the cross-view interaction with ego motions for better depth estimation. Specifically, we first calculate the corresponding sampling positions $P_{ij}^{source}$ in all source images at timestamp $t-1$ from the depth hypotheses $P_i^{ref} \in \mathbb{R}^{2 \times {C_D}' \times \frac{H}{n} \times \frac{W}{n}}$ in the $i$th reference image at timestamp $t$, using the following homography matrix:
\begin{equation} \label{temporal_stereo}
    H_{ij}(d)=K_j\cdot R_j\cdot(I-\frac{(T_i-T_j)\cdot n_{1}^{\top}}{d})\cdot R_{i}^{\top} \cdot K_{i}^{-1},
\end{equation}
where $H_{ij}(d)$ is the homography matrix between the $j$th source feature map at previous frame $t-1$ and the reference feature map of the $i$th image in frame $t$ at the sampled depth $d$. $T_j$ and $R_j$ are the translation and rotation matrix from the camera coordinate of the $j$th image at the timestamp $t-1$ to the ego coordinate of timestamp $t$. $T_i$ and $R_i$ are the translation and rotation matrices from the camera coordinate of the reference image to its ego coordinate at timestamp $t$. $K_j$ and $K_i$ are the camera’s intrinsic parameter of source image and reference image. $n_1$ denotes the principle axis of the reference camera. ${C_D}'$ is a reduced number of depth hypotheses for only the temporal depth estimation part to save the memory cost. Therefore, $P_{ij}^{source}$ is written as:

\begin{equation}
    P_{ij}^{source} = H_{ij}\cdot P_i^{ref}, \quad j=1,2...,N.
\end{equation}

Since a pixel in the reference image can only appear in part of source images, most coordinates in $P_{ij}^{source}$ is invalid. Therefore, we only sample the features in source frames with valid coordinates $\hat{P_{ij}^{source}}$. This is also the main difference between our STS and normal temporal stereo technique. As a result, we obtain the the warped feature volume $V_{i} \in \mathbb{R}^{{C_F} \times {C_D}' \times \frac{H}{n} \times \frac{W}{n}}$ using $\hat{P_{ij}^{source}}$.

Next, we adopt the group-wise correlation similarity measure~\cite{gwcnet} to build the lightweight cost volumes. Specifically, the feature channels of reference image feature map $F_{i}$ at time $t$ and the feature map $F_{v}$ from feature volume $V_{i}$ are firstly divided into $G$ groups and then the group-wise correlation for the $g$-th feature group is calculated as:
\begin{equation} \label{gwc}
    S_{i}^g=\frac{1}{{C_F}/G}\left \langle {F_i^{g}, F_v^{g}}\right \rangle,
\end{equation}
where $\left \langle \cdot ,\cdot\right \rangle$ is the inner product. All the $G$ group correlations are packed into a G-channel cost volume $C_{i}\in \mathbb{R}^{G \times {C_D}' \times\frac{H}{n}\times \frac{W}{n}}$.

We apply three 3D Convolutions with kernel size $1\times1\times1$ to regularize the cost volume, and an Average Pooling to adjust the shape of depth logits from $\frac{H}{n}\times \frac{W}{n}$ to $\frac{H}{m}\times \frac{W}{m}$. Moreover, we expand the number of depth channels ${C_D}'$ to ${C_D}$ by simply duplicating the origin depth feature values to get the final depth logits from STS $D_i^{stereo}\in \mathbb{R}^{{C_D} \times\frac{H}{m}\times \frac{W}{m}}$. As a result, the shape of depth from STS is aligned to monocular depth, facilitating the subsequent depth fusion technique. 

\paragraph{Depth Sampling}

Common depth estimation uses a uniform sampling method. However, it is not suitable for stereo-based depth estimation. After projecting the uniform sampled depth points to another view, the distance between neighbor points varies with the value of depth. Especially for the  points at a close range, their projected counterparts will be very sparse which will cause no points to fall in the correct location. As shown in Figure~\ref{fig:sid_ud_com}, it may fail to sample the corresponding point on the source frame, especially in closer objects in stereo-based depth estimation. In this case, stereo-based methods perform poorly. Thus we propose to use the depth discretization strategy, named Spacing-Increasing Discretization(SID)~\cite{dorn}:
\begin{equation} \label{sid_num}
    d = exp(\log(D_{min}) + \frac{\log(D_{max} / D_{min})*k)}{C_D},
\end{equation}

where $D_{min}$ and $D_{max}$ represent the start and end of the depth range. SID strategy samples depth in log space, which will benefit a lot to sample points at a close range for our surrounding-view temporal stereo module.  

\begin{table*}[!t]
\centering
\resizebox{\textwidth}{!}{
\begin{tabular}{c|c|c|cccccc|c}
\toprule
 \textbf{Method} & \textbf{Backbone} &\textbf{Resolution}  & \textbf{mAP}$\uparrow$  & \textbf{mATE}$\downarrow$ & \textbf{mASE}$\downarrow$  & \textbf{mAOE}$\downarrow$ & \textbf{mAVE}$\downarrow$ & \textbf{mAAE}$\downarrow$ & \textbf{NDS}$\uparrow$ \\
  \midrule
  BEVDepth &\multirow{2}{*}{ResNet-50}& \multirow{2}{*}{256$\times$704} & 0.351 & 0.639  &  0.267   & 0.479 & 0.428	 & 0.198 & 0.475   \\
  Ours & &  & 0.377 & 0.601  &  0.275 & 0.450 & 0.446 &	0.212 & 0.489  \\ 
  \midrule
  BEVDepth &\multirow{2}{*}{ResNet-50}& \multirow{2}{*}{512$\times$1408} & 0.405 & 0.570  &  	0.266   & 	0.383 & 0.368 & 0.206 & 0.523   \\
  Ours & &  & 0.425 & 0.532  &  0.267 & 0.390 & 	0.369 & 0.212 & 0.536  \\ 
  \midrule
  BEVDepth&\multirow{2}{*}{ConvNeXt}& \multirow{2}{*}{512$\times$1408} & 0.462  & 0.540 & 0.254	&  0.353 & 0.379 & 0.200	 & 	0.558  \\
  Ours & &  & 0.473 & 0.515  &  0.259 & 0.320 & 0.366 & 0.197 & 0.571  \\   
  \bottomrule
\end{tabular}}\caption{Comparison to BEVDepth~\cite{bevdepth} on the nuScenes \emph{val} set. CBGS is adopted for all experiments.}
\label{tab:compare_bevdepth}
\end{table*}

\subsection{Monocular Depth Learning and Depth Fusion}
Since temporal stereo can not handle texture-less areas and moving objects, we thus retain the monocular depth module in~\cite{bevdepth}. Following BEVDepth, our monocular depth module accepts features from backbone with down-sampling factor $m$ as its input, and then leverages a Camera-awareness Depth Estimation to predict monocular depth $D_i^{mono}\in \mathbb{R}^{{C_D} \times\frac{H}{m}\times \frac{W}{m}}$.

For depth fusion, we simply adopt a naive ``element-wise sum'' strategy since we find it simple but effective. Specifically, we add the monocular depth logits $D_i^{mono}$ and the temporal depth logits $D_i^{stereo}$ together and use the Softmax function $\sigma$ to obtain the final depth probability:
\begin{equation} \label{depth_fusion}
    D_i^{pred}=\sigma(D_i^{stereo}+D_i^{mono}).
\end{equation}

We project the point clouds to the image planes to obtain the ground-truth depth and use Binary Cross Entropy Loss for training depth modules.

\section{Experiment}
\subsection{Experimental Setup}
\subsubsection{Dataset}

We conduct experiments on public autonomous driving benchmark nuScenes~\cite{nuscenes}. The nuScenes dataset collects 1000 scenes using cameras, lidar, and radars. To be more specific, there are five cameras with 360 horizontal FOV, one lidar, and five radars. 1000 scenes are divided into 700/150/150 for training/validation/test. 10 classes are annotated for object detecting task including: car, truck, bus, trailer, construction vehicle, pedestrian, motorcycle, bicycle, barrier, and traffic cone.

We use official evaluation metrics to evaluate our detector, including mean Average Precision (mAP), mean Average Translation Error (mATE), mean Average Scale Error (mASE), mean Average Orientation Error (mAOE), mean Average Velocity Error (mAVE), mean Average Attribute Error (mAAE), and nuScenes Detection Score (NDS).

\subsubsection{Implementation details}
We use state-of-the-art in multi-view 3D detection BEVDepth~\cite{bevdepth} with ResNet-50 as our baseline model, and apply our STS on it. Following BEVDepth, we set 256 $ \times $ 704 as our basic experimental resolution and use the key and 4$th$ sweep images as input. $C_D$ is set to 112 by default. We also adopt data augmentations like image cropping, scaling, flipping, rotation, and BEV feature scaling, flipping, and rotation. We use AdamW with EMA as the optimizer. For the ablation study, we train the model with total batch size 64 without CBGS~\cite{cbgs} strategy for 24 epochs, while the benchmark results are trained with CBGS. The default features used for stereo matching and monocular depth learning is stride 16 and stride 4, respectively.

\subsection{Benchmark Results}
 We present results of BEVDepth with STS on nuScenes \emph{val} set. Table~\ref{tab:compare_bevdepth} shows the comparison between BEVDepth and STS with different settings. On the baseline setting, STS boosts BEVDepth in mAP, mATE, and NDS by 2.6\%, 0.038, and 1.4\%, respectively. We also study STS's effect when using a larger input resolution and a larger backbone. When input size is set to 512 $\times$ 1408, STS can still improve mAP and NDS by 2.0\% and 1.3\%. Moreover, on a much stronger backbone ConvNeXt, our STS still brings non-marginal improvement, demonstrating STS's effectiveness. We also compare STS with other leading 3D detection methods in Table~\ref{tab:results-detection}. Surprisingly, STS with a small backbone such as ResNet-50 even outperforms other methods in larger backbone like ResNet-101 or larger resolution like $900\times1600$. STS with R101 backbone surpasses existing methods by 1.5\% mAP and 0.7\% NDS. It boosts BEVFormer by 1.5\% on mAP and 2.5\% on NDS, and boosts BEVDet4D 3.5\% on mAP and 2.7\% on NDS. 
 
Besides, as a metrics designed to measure objects locations, mATE is highly related to depth estimation. When we consider mATE separately,  it is easy to find STS can bring a great boost. STS can outperform other methods in mATE by at least 3\% and bring improvement over 10\%  to methods except BEVDepth.

\begin{table*}
\centering
\resizebox{\textwidth}{!}{
\begin{tabular}{l|c|c|ccccc|c}
\toprule
 \textbf{Method} & \textbf{Resolution}  & \textbf{mAP}$\uparrow$  & \textbf{mATE}$\downarrow$ & \textbf{mASE}$\downarrow$  & \textbf{mAOE}$\downarrow$ & \textbf{mAVE}$\downarrow$ & \textbf{mAAE}$\downarrow$ & \textbf{NDS}$\uparrow$ \\
\midrule
FCOS3D~\cite{fcos3d} & 900$\times$1600 & 0.295 &0.806 &0.268 &0.511 &1.315 &0.170 &0.372 \\
DETR3D~\cite{detr3d} & 900$\times$1600 & 0.303 &0.860 &0.278 &0.437&0.967 &0.235 &0.374 \\

PETR-R50~\cite{petr} & 384$\times$1056 & 0.313 &0.768 &0.278 &0.564 &0.923 &0.225 &0.381 \\
PETR-R101 & 512$\times$1408 & 0.357 &0.710 &0.270 &0.490 &0.885 &0.224 &0.421 \\
PETR-Tiny& 512$\times$1408 & 0.361 &0.732 &0.273 &0.497 &0.808 &\textbf{0.185} &0.431 \\

BEVDet-Base & 512$\times$1408 & 0.349 &0.637 &0.269 &0.490 &0.914 &0.268 &0.417 \\
BEVDet4D-Base & 640$\times$1600 & 0.396 &0.619 &\textbf{0.260} &0.361 &0.399 &0.189 &0.515 \\

BEVFormer-S~\cite{bevformer}& - & 0.375 &0.725 &0.272 &0.391 &0.802 &0.200 &0.448 \\
BEVFormer-R101-DCN& 900$\times$1600 & 0.416 &0.673 &0.274 &0.372 &0.394 &0.198 &0.517 \\

BEVDepth-R101& 512$\times$1408 & 0.412 &0.565 &0.266 &\textbf{0.358} & \textbf{0.331} &0.190 &0.535 \\ \midrule
Ours-R50 & 512$\times$1408 & 0.425 & 0.532  &  0.267 & 0.400 & 0.369 &	0.212 & 0.536  \\ 
Ours-R101 & 512$\times$1408  & \textbf{0.431} & \textbf{0.525}  &  0.262 & 0.380 & 	0.369 & 0.204 & \textbf{0.542}  \\ 
\bottomrule
\end{tabular}}
\caption{Comparing STS to other state-of-the-art 3D detectors on the nuScenes \emph{val} set.}
\label{tab:results-detection}
\end{table*}

\subsection{Ablation Study}

To delve into the effect of different modules, we conduct several ablation experiments on nuScenes \emph{val} set.

\paragraph{Surround-view Stereo.} The main difference between Surround-view Temporal Stereo and normal Temporal Stereo is whether cross-camera interaction at different timestamps is allowed. When not using ``surround-view'', we only perform stereo matching on images from the same camera across time. If a resulting epipolar line falls out of that same camera (often happens in cameras on the left and right sides), the stereo matching procedure at that pixel is simply discarded. Results (see Table~\ref{tab:ablation-sts}) show that surround-view improves mAP by 1.1\% (35.6\% \emph{v.s.} 34.5\%), which almost contributes half of gain from the whole stereo mechanism. These results demonstrate the necessity of bringing ``surround-view'' into Temporal Stereo.

\begin{table}[t]
\centering
\begin{tabular}{c|c|cc|c}
\toprule
\textbf{Method}                & \textbf{Sur. view} & \textbf{mAP}$\uparrow$  & \textbf{mATE}$\downarrow$ & \textbf{NDS}$\uparrow$ \\ \midrule
BEVDepth              &     -         & 0.332    & 0.692 & 0.440    \\ \midrule
\multirow{2}{*}{Ours} &               &  0.345   & 0.663 & 0.451    \\
                      & \checkmark & \textbf{0.356} & \textbf{0.648}  &  \textbf{0.455}     \\ \bottomrule
\end{tabular}\caption{Ablation study on whether allowing stereo matching between across cameras (\emph{i.e.}, whether using surround-view). ``Sur. view'' denotes surround-view. CBGS is not used.}
\label{tab:ablation-sts}
\end{table}

\paragraph{Spacing-Increasing Discretization} SID facilitates the stereo module by making the projected sampling points on the source frame uniform (Figure~\ref{fig:sid_ud_com}). See Table~\ref{tab:ablation-sid}, using SID in STS improves mAP, mATE and NDS. However, using SID in the monocular depth branch in naive BEVDepth does not work. These results show that the benefit of SID fully comes from the improvement of temporal stereo depth estimation. Results in \ref{tab:different-bin} further reveal that SID mainly improves the mAP and mATE in short- and mid-range distances, which is consistent with our observations on the improvement of the sampling points in Figure~\ref{fig:sid_ud_com}.

\begin{table}[t]
\centering
\begin{tabular}{c|c|cc|c}
\toprule
 \textbf{Method} & \textbf{SID} & \textbf{mAP}$\uparrow$  & \textbf{mATE}$\downarrow$ & \textbf{NDS}$\uparrow$ \\
\midrule
  \multirow{2}{*}{BEVDepth} &  & 0.332 & 0.692	 & 	0.440 \\
    & \checkmark & 0.329 & 0.697 &  0.439  \\ \midrule
  \multirow{2}{*}{Ours}  &  & 0.350	 & 0.661  &  0.451  \\
    & \checkmark & \textbf{0.356} & \textbf{0.648}  &  \textbf{0.455}   \\ 
\bottomrule
\end{tabular}\caption{Ablation study on the effectiveness of SID in BEVDepth and BEVDepth with STS.}
\label{tab:ablation-sid}
\end{table}

\paragraph{Features for Stereo Matching} Features with different down-sampling factors contain different information. Low-level features preserve more texture and shape information, while high-level features are more semantically rich. We explore features with different down-sampling strides for STS. As shown in Table~\ref{tab:ablation-num_layers}, low-level features with high resolution can obtain better performance in terms of mAP, mATE and NDS. The performance could be further improved if we continuously increase the feature resolution. However, due to the huge memory cost on the 4D cost column, we leave it as a future work. 

\begin{table}[t]
\centering
\begin{tabular}{c|cc|c}
\toprule
 \textbf{D.S. Factor} & \textbf{mAP}$\uparrow$  & \textbf{mATE}$\downarrow$ & \textbf{NDS}$\uparrow$ \\
\midrule
  16 & 0.335 & 0.689  &  0.443   \\
  8 & 0.347 & 0.653 & 0.448    \\
  4 & \textbf{0.356} & \textbf{0.648} & \textbf{0.455}    \\
 \bottomrule
\end{tabular}
\caption{Ablation Study of features used for stereo matching. D.S Factor stands for down-sampling factor.}
\label{tab:ablation-num_layers}
\end{table}

\paragraph{Number of Depth Bins} We further explore the effect of the number of depth bins ${C_D}'$ in Table~\ref{tab:ablation-num_points}. Note that ${C_D}'$ is not the same as $C_D$ since $C_D$ is the number of final depth bins before feature unprojection. As we stated in Section~\ref{method}, the number of depth bins used for STS is equal to or less than $C_D$ to save the huge memory it costs. The final depth logits are duplicated to match the final $C_D$. Table~\ref{tab:ablation-num_points} shows that as ${C_D}'$ increases, the performance is improved. But even with the least number of depth bins 7, mAP is still improved. The improvement carried by this factor saturates when it is greater than 56. We thus set it as 56 for STS in terms of the balance of accuracy and memory cost. 

\begin{table}[t]
\centering
\begin{tabular}{c|c|cc|c}
\toprule
 \textbf{Method} & \textbf{${C_D}'$} & \textbf{mAP}$\uparrow$  & \textbf{mATE}$\downarrow$ & \textbf{NDS}$\uparrow$ \\
\midrule
  BEVDepth & - & 0.332 & 0.692 & 0.440   \\ \midrule
  
   \multirow{5}{*}{Ours} & 7  & 0.339 & 0.684  & 0.441 \\ 
   & 14 & 0.341 & 0.685 & 0.444  \\
   & 28 & 0.352 & 0.673 & 0.454   \\
   & 56 & 0.356 & 0.648 &  0.455  \\
   & 112 & \textbf{0.357} & \textbf{0.643} &  \textbf{0.458}  \\
\bottomrule
\end{tabular}\caption{Ablation study on ${C_D}'$ in STS.}
\label{tab:ablation-num_points}
\end{table}

\begin{table}[t]
\centering
\begin{tabular}{c|c|cc|c}
\toprule
 \textbf{Monocular} & \textbf{Stereo} & \textbf{mAP}$\uparrow$  & \textbf{mATE}$\downarrow$ & \textbf{NDS}$\uparrow$ \\
\midrule
  \checkmark & & 0.334 & 0.700 & 0.442   \\
  & \checkmark & 0.312  & 0.737 & 0.412   \\
  \checkmark &\checkmark & \textbf{0.356} & \textbf{0.648} & \textbf{0.455}  \\
\bottomrule
\end{tabular}\caption{Ablation study on solely using monocular depth branch and STS depth branch.}
\label{tab:ablation-mono-stereo}
\end{table}

\paragraph{Monocular and temporal stereo fusion} In this part, we study the effect of depth fusion. Specifically, we train BEVDepth with STS, and test only using one depth module, either depth from STS or monocular depth branch. As shown in Table~\ref{tab:ablation-mono-stereo}, compared with the one only using the monocular module, fusing the depth from monocular and STS boosts the performance in mAP and NDS by about 2.2\% and 1.3\%. mATE is reduced from 0.70 to 0.648. Compared with the one only using the depth from STS module, fusing the two kinds of depth together boosts the performance in mAP by about 4.4\%. The performance gain on depth fusion shows that monocular depth and depth from STS both have their own advantages, which can be simultaneously leveraged by fusing them.

\subsection{Further Analysis}

We conduct qualitative and quantitative analysis to study how STS improves depth estimation and 3D detection. 

\begin{figure}[!t]
\centering
\includegraphics[width=0.46\textwidth]{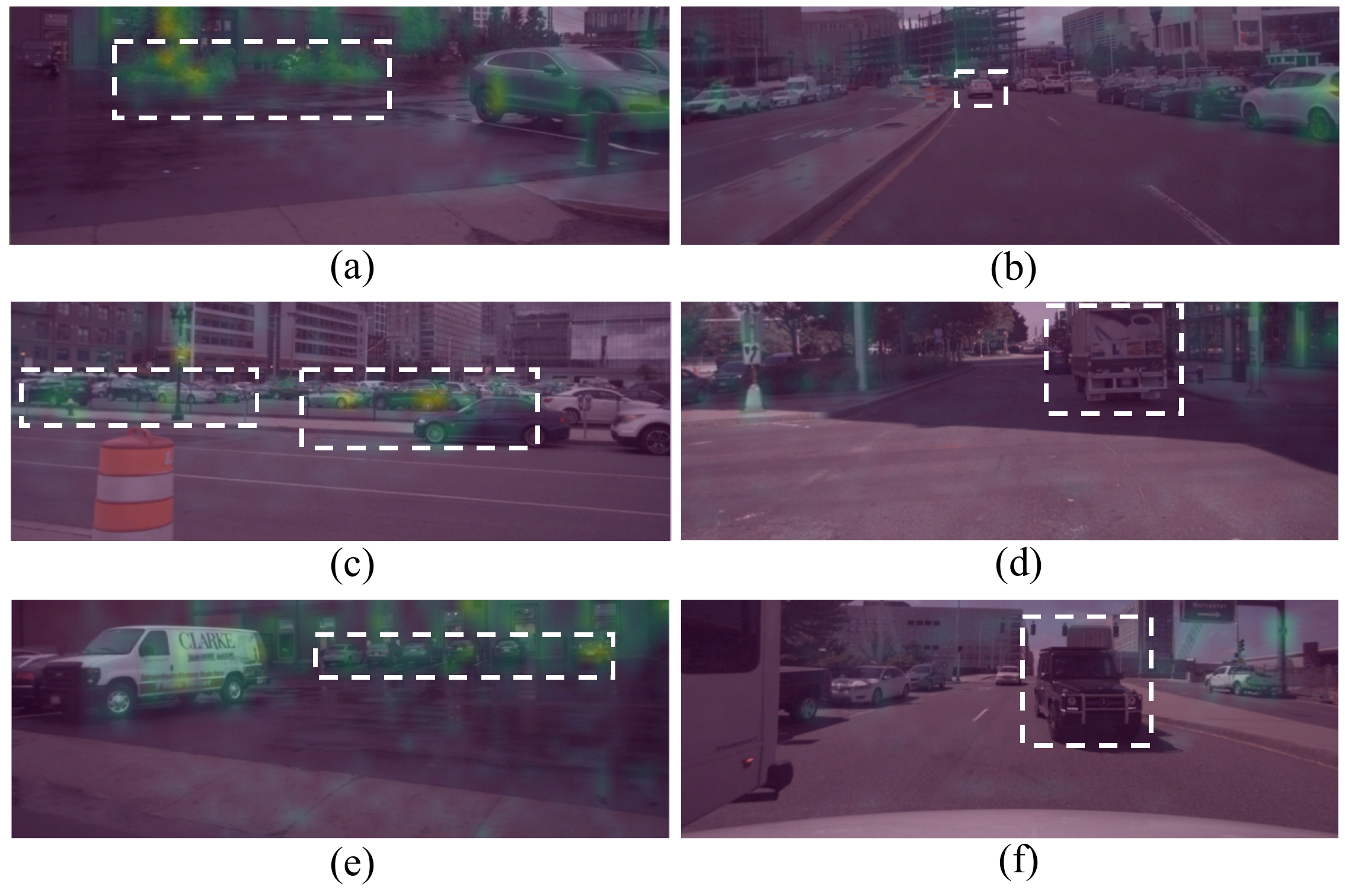} 
\caption{Visualization of regions that STS has positive effects on. Dashed boxes highlight key regions for analysis.}
\label{fig:weight}
\end{figure}

\paragraph{Visualization} We visualize depth results improved by STS. Specifically, we calculate $\sigma(D^{stereo}+D^{mono}) - \sigma(D^{mono})$ and show only its positive values at the correct depth bin in Figure~\ref{fig:weight}. (a) (c) and (e) show that STS improves depth mostly at medium and long distances while depth value at a short distance such as on the ground is barely affected; Objects in dashed boxes in (b) (d) and (f) represent moving objects, depth of which is not improved at all. Moreover, in (a)-(e), depth in texture-less regions such as the ground, surface of buildings and walls is not improved by STS. The phenomenon of moving objects and texture-less regions is consistent with the consensus in MVS.

\paragraph{Quantitative Analysis} We evaluate 3D detection performance in separated ranges according to objects' distances to the ego vehicle, including 2-10 meters, 10-20 meters, 20-30 meters, 30-45 meters, and 45-58 meters. We can see from Table~\ref{tab:different-bin} that mAP is barely improved in short distances. It even drops within 2-20 meters when not using SID, indicating that the monocular depth estimation within this range is fairly reliable. As distance increases, the model with STS performs better than the model with only monocular depth, demonstrating that STS compensates for the monocular depth estimation in the far distance. This observation is consistent with the visualization results in Figure~\ref{fig:weight}. Moreover, as we mentioned in Section~\ref{method}, SID helps sample features in source frames more uniformly, see detection performance in Figure~\ref{fig:weight}, especially within 2-10 and 21-30 meters. The improvement within these ranges clearly shows the importance of non-uniform depth hypotheses, which is consistent with the observation~\cite{xu2020learning} in MVS.

\begin{figure}[!t]
\centering
\includegraphics[width=0.46\textwidth]{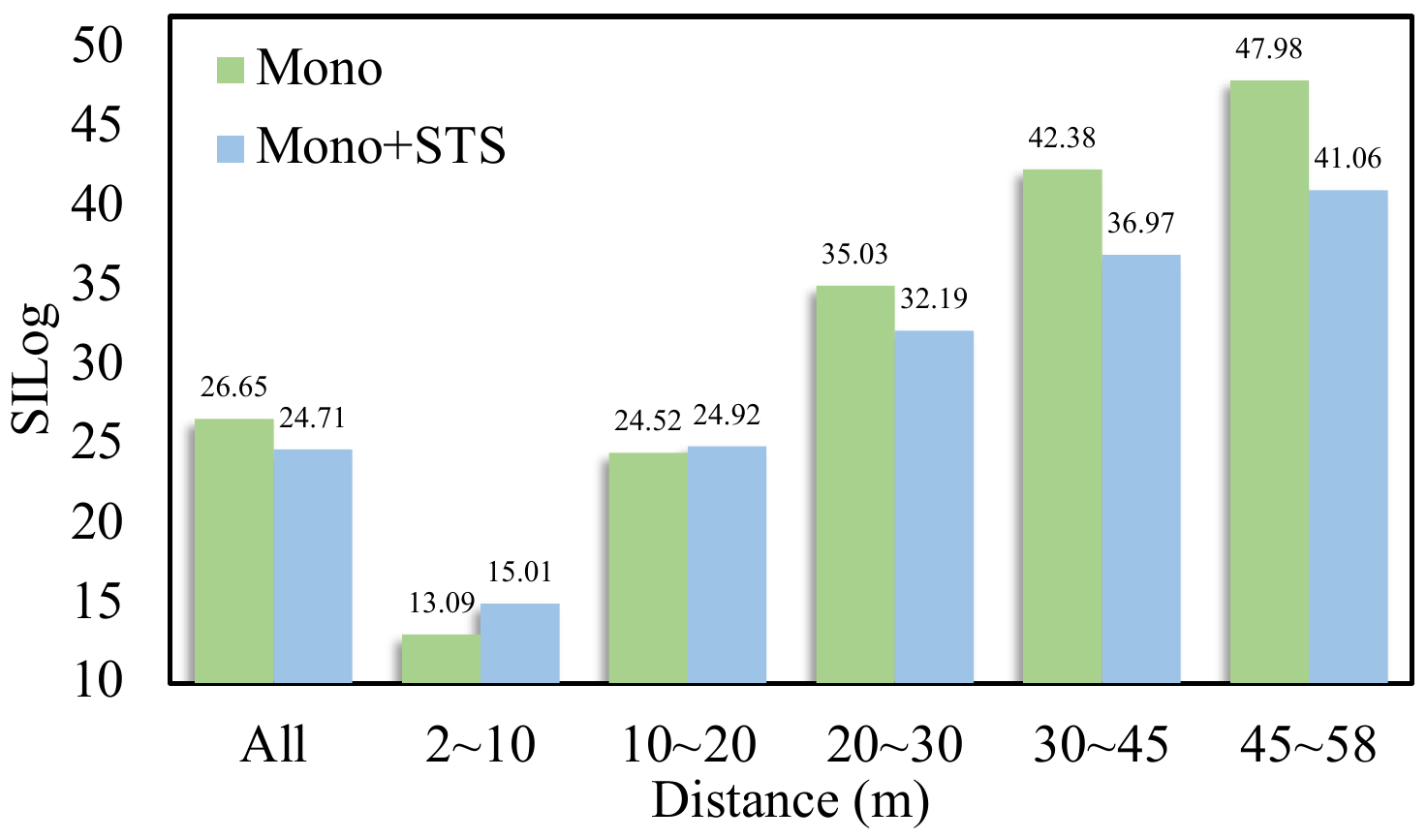} 
\caption{Comparing depth between monocular depth (BEVDepth) and depth learning with STS. Depth is evaluated in different distance ranges using SILog error.}
\label{fig:depth}
\end{figure}

\paragraph{Depth Evaluation} Following 3D detection, we evaluate depth in different ranges using SILog error. See Figure~\ref{fig:depth}, although depth error in short distances slightly increases, it greatly reduces in long distance, especially within 30-45 and 45-58 meters, SILog of which reduces 5.41 and 6.92, respectively. These results demonstrate again that STS mainly facilitates depth learning in mid- and long-range areas.

\begin{table}[!t]
\centering
\resizebox{0.48\textwidth}{!}{
\begin{tabular}{c|cc|cc|c}
\toprule
 \textbf{Distance (m)} & \textbf{STS} & \textbf{SID} & \textbf{mAP}$\uparrow$  & \textbf{mATE}$\downarrow$ & \textbf{NDS}$\uparrow$  \\
\midrule
 \multirow{4}*{2-10} & &  &0.622  &0.339 &0.631   \\
  &\checkmark &  &0.617	& 0.327 & 0.628     \\
  &\checkmark & \checkmark & \textbf{0.626}  & \textbf{0.319} &\textbf{0.646}  \\
  
  \midrule
 \multirow{4}*{11-20} &  &  & \textbf{0.510} & 0.536 & \textbf{0.577}  \\
  &\checkmark &  &0.500 & \textbf{0.508} & 0.566 \\
  &\checkmark & \checkmark  &0.508 &0.519 & 0.572  \\

  \midrule
 \multirow{4}*{21-30} &  &  & 0.252  &0.787	&0.418  \\
  &\checkmark &  & 0.272 &0.757 &0.422 \\
  &\checkmark & \checkmark & \textbf{0.283} & \textbf{0.727} & \textbf{0.439} \\

  \midrule
 \multirow{4}*{31-45} &  &  &0.096  &0.961	& 0.272	\\
  &\checkmark &  &\textbf{0.099} &0.942 &\textbf{0.278}  \\
  &\checkmark & \checkmark  &0.096 & \textbf{0.924} & 0.276  \\

  \midrule
 \multirow{4}*{46-58} &  &  &0.018  &0.991 &0.113  \\
  &\checkmark &  &\textbf{0.020} & \textbf{0.983}  & \textbf{0.115}  \\
  &\checkmark & \checkmark  &0.019 &1.001  &0.109 \\
\bottomrule
\end{tabular}}
\caption{Performance comparison on the nuScenes \emph{val} set with objects in different distances.}\label{tab:different-bin}
\end{table}

\section{Conclusion}
In this paper, we introduce Surround-view Temporal Stereo (STS) for multi-view 3D object detection. The core insight of STS is to utilize the temporal stereo technique in multi-view 3D object detection by regarding field of views from all cameras as a unified view. We also leverage a non-uniform depth hypotheses to better utilizing the geometry correspondence to learn the real-world scale depth. Moreover, we incorporate a monocular stereo depth fusion mechanism for the final dense depth distribution the moving objects and texture-less areas. Experimental results on nuScenes demonstrate its effectiveness for multi-view 3D detection.

\bibliography{aaai23}

\end{document}